\documentclass[conference]{IEEEtran}
\IEEEoverridecommandlockouts
\usepackage{cite}
\usepackage{amsmath,amssymb,amsfonts}
\usepackage{algorithmic}
\usepackage{graphicx}
\usepackage{textcomp}
\usepackage{xcolor}
\def\BibTeX{{\rm B\kern-.05em{\sc i\kern-.025em b}\kern-.08em
    T\kern-.1667em\lower.7ex\hbox{E}\kern-.125emX}}
\begin{document}

\title{Depth-Enhanced YOLO–SAM2 Detection for Reliable Ballast Insufficiency Identification}

\author{
    Shiyu Liu\textsuperscript{1}, 
    Dylan Lester\textsuperscript{1}, 
    Husnu Narman\textsuperscript{2}, 
    Ammar Alzarrad\textsuperscript{3}, 
    and Pingping Zhu\textsuperscript{1*},
    
    \thanks{This research was supported by the US Army Corps of Engineers (USACE) - The Engineer Research and Development Center (ERDC) - Grant: W912HZ249C006 }
    
    \thanks{
        \textsuperscript{1}Shiyu Liu, Dylan Lester, and Pingping Zhu are with the Department of Biomedical and Electrical Engineering, Marshall University, Huntington, WV 25755, USA. 
        (e-mails: liu137@marshall.edu; lester299@marshall.edu; zhup@marshall.edu)
    }
    
    \thanks{
    	\textsuperscript{2}Husnu Narman is with the Department of Computer Science, Marshall University, Huntington, WV 25755, USA. 
    	(e-mails: narman@marshall.edu)
    }
    
    \thanks{
    	\textsuperscript{3}Ammar Alzarrad is with the Department of Civil  Engineering, Marshall University, Huntington, WV 25755, USA. 
    	(e-mails: alzarrad@marshall.edu)
    }
    
    \thanks{
        *Corresponding author: Pingping Zhu (email: zhup@marshall.edu)
    }
}

\maketitle

\begin{abstract}
This paper presents a depth-enhanced YOLO–SAM2 framework for detecting ballast insufficiency in railway tracks using RGB–D data. Although YOLOv8 provides reliable localization, the RGB-only model shows limited safety performance, achieving high precision (0.99) but low recall (0.49) due to insufficient ballast, as it tends to over-predict the sufficient class. To improve reliability, we incorporate depth-based geometric analysis enabled by a sleeper-aligned depth-correction pipeline that compensates for RealSense spatial distortion using polynomial modeling, RANSAC, and temporal smoothing. SAM2 segmentation further refines region-of-interest masks, enabling accurate extraction of sleeper and ballast profiles for geometric classification.

Experiments on field-collected top-down RGB–D data show that depth-enhanced configurations substantially improve the detection of insufficient ballast. Depending on bounding-box sampling (AABB or RBB) and geometric criteria, recall increases from 0.49 to as high as 0.80, and F1-score improves from 0.66 to over 0.80. These results demonstrate that integrating depth correction with YOLO–SAM2 yields a more robust and reliable approach for automated railway ballast inspection, particularly in visually ambiguous or safety-critical scenarios.
\end{abstract}


\section{Introduction}
Railway ballast is a foundational material that's placed beneath and between railroad ties to support track infrastructure. It plays a crucial role in distributing the weight of trains, holding the tracks in place, and allowing for proper drainage. Maintaining sufficient ballast is essential to ensure the structural integrity and operational safety of railway systems.

However, traditional ballast inspection methods rely on manual visual inspection. It is labor-intensive, inconsistent due to subjective judgment across different workers and environmental conditions, and sometimes unsafe, as inspectors must physically access active track environments.

Recent advances in computer vision have enabled automated analysis of railway imagery\cite{b7},\cite{b8},\cite{b9}. YOLO (You Only Look Once)\cite{b1},is a real-time object detection framework that rapidly predicts the locations and classes of objects within an image using a single neural network. YOLO-based models can efficiently identify and localize ballast regions from RGB images. However, relying only on RGB data is insufficient. It can't capture the depth information necessary to determine whether the ballast is physically sufficient.

Depth sensing offers geometric information that can address these limitations. The integration of RGB-D sensing has gained traction in broader infrastructure inspection tasks\cite{b10},\cite{b11},\cite{b12}. However, depth sensors such as the Intel RealSense often show spatially varying bias and surface warping due to sensor tilt and environmental conditions. Without correcting these depth distortions, the measurements are unreliable. Many researchers have investigated distortion-correction techniques for RealSense depth sensors\cite{b13},\cite{b14},\cite{b15}.

The broader aim of this paper is to enhance the safety and reliability of railway operations by enabling accurate, automated, and repeatable assessment of ballast conditions using vision-based methods. The specific objective of this work is to develop an RGB–D inspection system that automatically detects insufficient ballast by integrating deep-learning-based region detection, precise segmentation, and geometrically robust depth correction. Our pipeline uses YOLOv8 \cite{b2} for initial ballast detection, followed by Segment Anything Model 2 (SAM2) \cite{b3} for precise mask refinement and rotated bounding-box extraction, ensuring accurate alignment with the physical orientation of the ballast regions. We introduce a novel depth-bias correction method that uses RANSAC-based \cite{b4} polynomial fitting and temporal smoothing to stabilize depth estimates over time, mitigating spatial distortions inherent in raw sensor data. Using this corrected depth data, we reconstruct ideal ballast planes within each rotated region and apply a dual-metric classification strategy to distinguish sufficient from insufficient ballast regions.

The effectiveness of this approach is validated in Section~\ref{sec:experiments}, where we compare three models: YOLO-only, YOLO with depth, and our rotated-box depth-enhanced algorithm. While YOLO-only achieves an F1-score of 0.74 but suffers from dangerous false positives, our rotated-box depth method achieves the best performance with an F1-score of 0.81 and a recall of 0.90 for insufficient ballast, representing a substantial improvement in safety-critical detection.

The remainder of this paper is organized as follows. Section II formulates the RGB--D ballast sufficiency detection problem. Section III presents the proposed methodology, including YOLO-based ballast detection, SAM2 segmentation with rotated bounding boxes, depth correction, and dual-criteria sufficiency classification. Section IV describes the dataset, evaluation metrics, and experimental results comparing the proposed approach with RGB-only and non-rotated depth baselines. Finally, Section V concludes the paper and discusses directions for future work.

The main contributions of this work are:

\begin{itemize}
    \item An integrated RGB-D pipeline for automated ballast inspection, combining YOLO detection and SAM2 segmentation, and rotated bounding-box extraction tailored to railway geometry.

    \item A robust spatial bias correction method using RANSAC-fitted polynomial surfaces and temporal smoothing, enabling consistent depth measurement without external calibration.

    \item A dual-criteria ballast sufficiency classifier that jointly evaluates global depth residuals and localized gap indicators to reliably identify insufficient ballast with high precision.
\end{itemize}

\section{Problem Formulation}
\label{Sec:Intro}
\begin{figure}
	\centering
	\includegraphics[width=\linewidth]{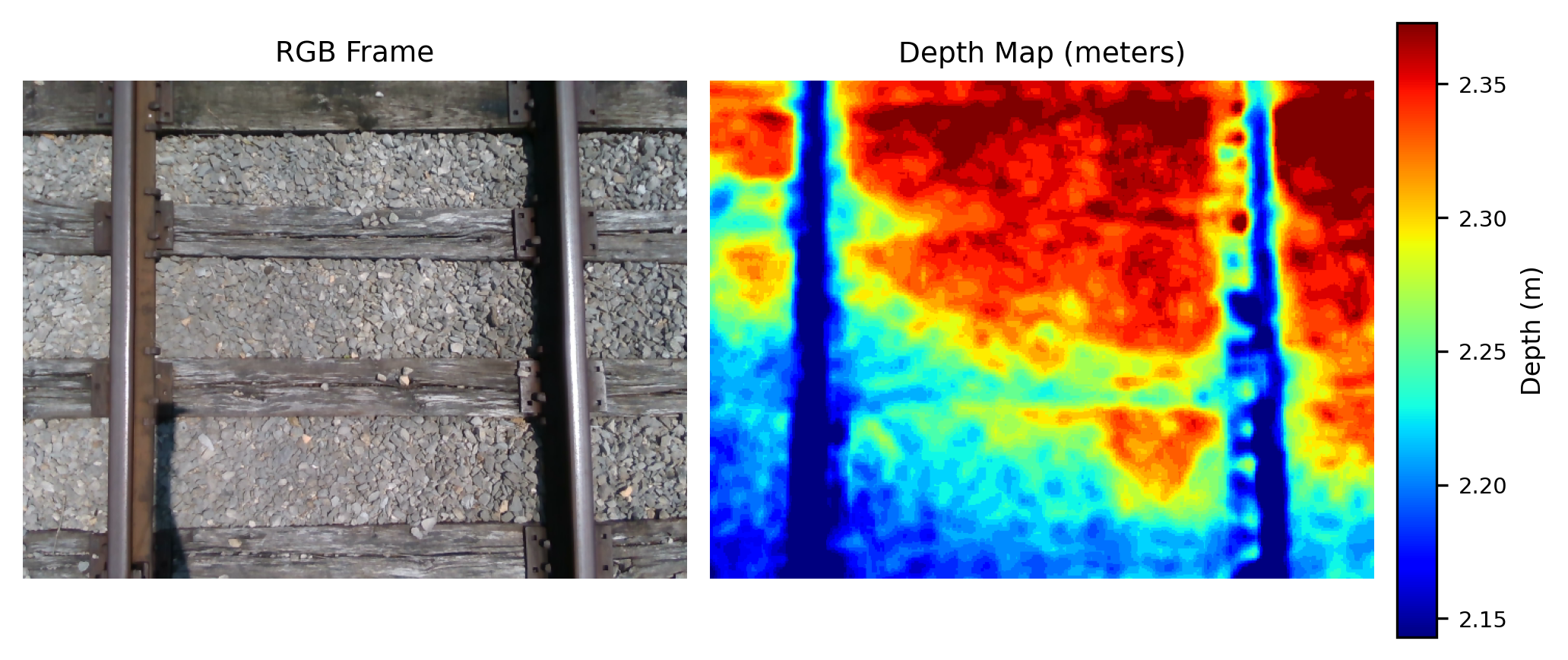}
	\caption{RGB frame (left) and corresponding raw depth map (right) captured by the RealSense D435.}
	\label{fig:overview}
\end{figure}

The goal of this project is to determine whether each sleeper bay contains sufficient ballast using the RGB and depth (RGB-D) image data captured by a top-mounted Intel RealSense D435 camera. As shown in Fig. \ref{fig:overview}, each frame consists of an RGB image represented by the function
\begin{equation}
 I_{\text{RGB}} :\Omega\rightarrow [0,255]^3,
\end{equation}
and a corresponding aligned depth map represented by
\begin{equation}
D_{\text{raw}}:\Omega\rightarrow \mathbb{R}^{+}.
\end{equation}
The shared domain of these functions is the discrete pixel grid  $(x,y)\in\Omega=[1,\ldots,W]\times[1,\ldots,H]$, which specifies the spatial coordinates of the RGB-D image pair. 

Following object detection, let $N$ denote the number of ballast segments identified in a given frame. The objective is to assign each segment a sufficiency label
\begin{equation}
s_i \in \{\text{sufficient}, \text{insufficient}\},
\qquad i = 1,\ldots,N
\end{equation}

In this paper, we assume that the RGB–D alignment provided by the \textit{Intel RealSense} software is accurate and dependable. Even under this assumption, two major challenges remain. \textbf{Challenge-I:} \textit{ballast segments appear with different orientations due to track geometry and the camera’s viewpoint, making rotated bounding regions necessary for consistent and precise sampling.} \textbf{Challenge-II:} \textit{the raw depth data exhibits spatially varying tilt and bias, which must be corrected before any reliable geometric analysis can be performed.}


\section{Methodology}
\begin{figure}
	\centering
	\includegraphics[width=1\linewidth]{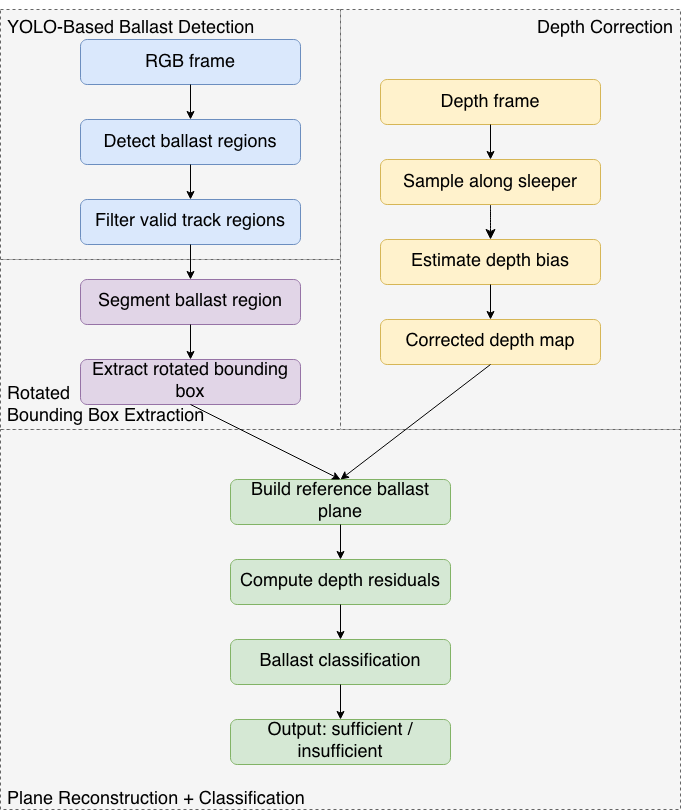}
	\caption{Flowchart of the proposed ballast sufficiency detection algorithm based on RGB–D sensing and rotated bounding box geometry}
	\label{flowchart}
\end{figure}

To achieve automated identification of insufficient ballast regions from RGB–D imagery, we introduce a novel framework that integrates YOLO-based detection with SAM2 segmentation, referred to as the YOLO–SAM2 approach. The overall architecture of the proposed method is illustrated in Fig. \ref{flowchart} and consists of four main components: (1) a YOLO-based ballast detection module, (2) a SAM2 segmentation and rotated bounding box extraction module, (3) a depth correction module, and (4) a plane reconstruction and ballast sufficiency classification module.  

\subsection{YOLO-Based Ballast Detection}\label{AA}
Our system first applies YOLOv8 to the RGB frame to localize ballast regions. Given an RGB image $I_{\text{RGB}}$, the existing YOLOv8 approach can predict a finite set of axis-aligned bounding boxes:
\begin{equation}
\mathcal{B}
=
\left\{
\boldsymbol{b}_i = (x_i, y_i, w_i, h_i, c_i)
\right\}_{i=1}^{N},
\label{eq:yolo_boxes}
\end{equation}
where $(x_i, y_i) \in \Omega$ are the center coordinates, $(w_i, h_i)$ denote the width and height of the $i$-th bounding box, and $c_i \in [0,1]$ is the corresponding
confidence score. After non-maximum suppression, low-confidence boxes are removed using a threshold $T_c$, such that $c_i \geq T_c$ for all $i\in \{1,\ldots, N\}$. 

Because insufficient ballast is typically found in the area between the rails, the YOLO-based detection is limited to the central 70\% of the image width. This constraint helps reduce false positives and ensures the analysis focuses on the relevant track region. The resulting bounding boxes within this area are then used as preliminary regions of interest (ROI) for subsequent segmentation and depth-based evaluation.

\subsection{SAM2 Segmentation and Rotated Bounding Box Extraction}\label{subsec: RBB}
Although YOLOv8 provides reliable initial localization, its axis-aligned bounding boxes often fail to match the actual geometric orientation of the ballast structures (\textbf{Challenge-I} in Section \ref{Sec:Intro}). To address this, each ROI is processed by the SAM2 segmentation module for mask refinement and rotated bounding-box extraction, ensuring alignment with the railroads' true physical layout.

Following the YOLO-based detection, each axis-aligned bounding box is refined using SAM2. Instead of generating a mask on the full image domain $\Omega$, we first crop the sub-image defined by the YOLO ROI $\boldsymbol{b}_i$ and use this cropped region as both the SAM2 input and the spatial prompt. This ROI-guided segmentation restricts SAM2 to the ballast instance identified by YOLO and prevents unrelated track elements from being included in the mask.

Formally, let $\Omega_i \subset \Omega$ denote the cropped region corresponding to $\boldsymbol{b}_i$. SAM2 produces a binary mask:
\begin{equation}
M_i(x,y) \in \{0,1\}, \qquad (x,y)\in \Omega_i,
\end{equation}
which is then mapped back to the full-image coordinates. Light post-processing, including morphological closing and removal of small isolated components, is applied to enhance mask coherence. Importantly, this procedure is applied independently to each preliminary ROI, resulting in one refined mask per YOLO detection.

To better align with the physical orientation of the ballast structures, we compute a rotated minimum-area bounding rectangle for each cleaned mask $M_i$:
\begin{equation}
\boldsymbol{b}_i^{\,r} = (x_i^{r}, y_i^{r}, \alpha_i, w_i^{r}, h_i^{r}),
\label{eq:rotated_box}
\end{equation}
where $(x_i^{r}, y_i^{r})$ denote the center of the rotated rectangular bounding-box in image coordinates, $\alpha_i \in (-\pi/2, \pi/2)$ is the rotation angle, and $w_i^{r}, h_i^{r}$ represent the width and height along the rotated axes. This rotated bounding box $\boldsymbol{b}_i^{\,r}$ is generated for each preliminary ROI, ensuring that every detected ballast region obtains its own geometry-aligned representation.

These rotated bounding boxes provide a spatially accurate representation of the ballast regions that aligns with the track orientation and enables more consistent depth sampling, which is critical for reliable depth correction and subsequent geometric analysis.

\subsection{Depth Correction}\label{AA}

Depth measurements obtained from \textit{Intel RealSense} sensors often exhibit systematic spatial biases caused by factors such as sensor tilt, lens distortion, and environmental conditions (\textbf{Challenge-II} in Section \ref{Sec:Intro}). These biases typically manifest as large-scale planar tilting or curved warping across the depth image, which must be corrected to ensure accurate ballast depth estimation. In this section, we present a robust depth correction method that models and removes these distortions using a polynomial surface fitted from sleeper-aligned depth samples.

\subsubsection{Polynomial Bias Model}\label{AA}

To compensate for the smooth, spatial distortion present in raw RealSense depth maps, we model the depth bias using a low-order 2D polynomial surface:
\begin{equation}
\Delta z(x,y \mid \boldsymbol{\theta})
=
\theta_1 x
+ \theta_2 y
+ \theta_3 x^2
+ \theta_4 y^2
+ \theta_5 x y
+ \theta_6,
\label{eq:bias_poly}
\end{equation}
where $(x,y)\in\Omega$ are pixel coordinates and $\boldsymbol{\theta}=(\theta_1,\theta_2,\theta_3,\theta_4,\theta_5,\theta_6)$ are the distortion parameters. The first five terms describe the spatially varying tilt–curvature field, while $\theta_6$ represents a global depth offset. Since ballast sufficiency depends only on relative height between ballast and sleepers, this constant offset does not influence the geometric decision.

With this decomposition, the raw depth measurement at pixel $(x,y)$ can be expressed as
\begin{equation}
D_{\mathrm{raw}}(x,y)
=
D_{\mathrm{true}}(x,y)
+
\Delta z(x,y \mid \boldsymbol{\theta})
+
\epsilon(x,y),
\label{eq:raw_depth_model}
\end{equation}
where $D_{\mathrm{true}}(x,y)$ is the unknown true depth and $\epsilon(x,y)$ represents high-frequency noise and residual fitting error. Then, we can have the corrected depth $D_{\mathrm{corr}}$, which is expressed by
\begin{equation}
	D_{\mathrm{corr}}(x,y\mid \boldsymbol{\theta}) = D_{\mathrm{raw}}(x,y) - \Delta z(x,y \mid \boldsymbol{\theta}) 
\end{equation}
The correct depth $D_{\mathrm{corr}}$ is an approximate of the unknown true depth $D_{\mathrm{true}}(x,y)$. 


This polynomial model provides a compact and effective approximation of the low-frequency distortion field. Subtracting the estimated bias surface removes large-scale tilt and curvature while preserving the geometric detail necessary for accurate ballast surface reconstruction and sufficiency classification.

\subsubsection{Sleeper Sample Extraction}\label{AA}

Accurate depth bias estimation relies on sampling from regions where the true depth profile follows a simple planar structure. In railway scenes, the sleeper surfaces between adjacent ballast segments meet this criterion. Therefore, we extract depth samples exclusively from sleeper regions to serve as reliable ground truth for polynomial bias fitting.

For each pair of neighboring rotated bounding boxes (rboxes) $\boldsymbol{b}_i^{r}$ and $\boldsymbol{b}_{i+1}^{r}$, we compute the midline between their adjacent edges as the sleeper sampling line. This midline is aligned with the orientation of the rboxes and ensures that depth samples lie precisely on top of the sleeper surface, avoiding contamination from ballast or background regions.

In cases where a ballast region lies at the top or bottom of the frame, without an adjacent rbox above or below. We apply a fallback strategy, extract depth samples along a horizontal line offset by $\Delta w = 10$ pixels above or below the rbox boundary, respectively. This ensures sampling consistency across frames and layout variations.

After extraction, raw depth values may still contain spikes or missing data due to sensor noise or segmentation imperfections. To enhance robustness, we apply Median Absolute Deviation (MAD) \cite{b5} filtering. Let $\mathcal{S}_d = \{(x_n, y_n, z_n)\}_{n=1}^{N_s}$ denote the set of sleeper-aligned raw depth samples, where $N_s$ is the total number of extracted samples and $z_n = D_{\mathrm{raw}}(x_n,y_n)$ is the raw depth at pixel $(x_n, y_n)$. The filtered sleeper depth samples are:
\begin{equation}
\mathcal{S}_d^{\mathrm{f}}
=
\left\{
(x_n, y_n, z_n) \in \mathcal{S}_d :
|z_n - \tilde{z}| < \tau \cdot \mathrm{MAD}(\mathcal{Z}_d)
\right\},
\label{eq:filtered_samples}
\end{equation}
where $\tilde{z}$ is the median of $\mathcal{Z}_d = \{z_n\}_{n=1}^{N_s}$, $\mathrm{MAD}(\mathcal{Z}_d)$ indicates the MAD of the set $\mathcal{Z}_d$,  and $\tau$ is a scalar threshold that determines the acceptable deviation range. Samples satisfying this condition are retained for subsequent bias surface estimation. This sleeper sampling strategy ensures that bias estimation operates on geometrically meaningful and noise-resistant depth measurements.

\subsubsection{RANSAC-Based Robust Bias Estimation}\label{AA}

The raw \textit{RealSense} depth measurements contain spatially varying bias caused by sensor tilt, stereo-matching distortion, and environmental noise. Since no ground-truth depth is available, we exploit the geometric fact that sleeper surfaces are approximately planar in the real world. Therefore, the raw depth samples extracted from sleeper regions, denoted by $\mathcal{S}_d^{\mathrm{f}}$, should ideally lie on a single smooth surface. Deviations within this set arise from sensor noise, segmentation imperfections, and partial occlusion.

To robustly estimate the polynomial bias surface under such outliers, we use the RANSAC~\cite{b4} procedure. In each iteration, a minimal subset of sleeper samples is randomly selected to generate a candidate bias model $\Delta z(x,y|\boldsymbol{\theta})$. All remaining samples are evaluated against this candidate, and those whose depth discrepancies fall below a tolerance threshold $T_{\mathrm{res}}$ are counted as inliers. The model producing the largest inlier set is selected as the best hypothesis.

Let $\mathcal{S}_{\mathrm{in}}$ denote the resulting inlier set. Each inlier sample $(x_n, y_n, z_n) \in \mathcal{S}_{\mathrm{in}}$ corresponds directly to a depth value $z_j $ taken from the raw depth map within the sleeper region. Because the true sleeper surface is approximately planar in the real world, any smooth, spatially coherent variation observed across these inlier depths cannot be explained by actual geometry. Instead, such low-frequency variation reflects the underlying distortion introduced by the depth sensor.

Therefore, the bias surface can be refined by minimizing  the approximated variance of the corrected depth data:
\begin{equation}
\boldsymbol{\theta}^*
 = \arg\min_{\boldsymbol{\theta}}
\sum_{(x_n,y_n,z_n)\in\mathcal{S}_{\mathrm{in}}}
\left[
D_{\mathrm{corr}}(x_n,y_n\mid \boldsymbol{\theta}) - \bar{z}_{\mathrm{in}}
\right]^2
\end{equation}
where $\bar{z}_{\mathrm{in}}$ is the average of $D_{\mathrm{corr}}(x_n,y_n | \boldsymbol{\theta})$ for all $(x_n,y_n,z_n)\in\mathcal{S}_{\mathrm{in}}$. Furthermore, considering the flat nature of sleepers, we assume that $\bar{z}_{\mathrm{in}}$ is a constant, and can be absorbed by the parameter $\theta_6$.  


\subsubsection{Depth Correction}\label{AA}

For each input frame, the polynomial bias parameters $\boldsymbol{\theta}^*$ estimated by RANSAC are used to correct the spatial depth distortion. To ensure smooth temporal transitions and reduce frame-to-frame flicker, we apply an exponential moving average (EMA) \cite{b5} filter to the bias coefficients. Let $\boldsymbol{\theta}_k^{\mathrm{raw}}$ denote the RANSAC-refined parameters from the $k$th frame (i.e., $\boldsymbol{\theta}_k^{\mathrm{raw}}=\boldsymbol{\theta}^*_k$ for
the $k$th frame). The temporally stabilized parameters are then computed as:
\begin{equation}
\boldsymbol{\theta}_k
=
\lambda\,\boldsymbol{\theta}_k^{\mathrm{raw}}
+
(1-\lambda)\,\boldsymbol{\theta}_{k-1},
\end{equation}
where $\boldsymbol{\theta}_k$ is the smoothed parameter vector applied to the
current frame, and $\lambda\in(0,1)$ is a smoothing factor that controls the relative contribution of the current vs. previous estimates.

This temporal stabilization ensures smooth transitions in the corrected depth field, minimizing fluctuations due to transient noise or unstable sampling regions.

Depth correction is then performed by removing only the spatially varying component of the estimated distortion field (the terms associated with $\theta_1$ through $\theta_5$). The constant offset parameter $c$ is not removed during correction and is intentionally preserved. Retaining this global offset keeps the corrected depth values within a numerically stable range and prevents the depth map from collapsing toward zero or negative values. Since ballast sufficiency depends solely on relative height differences between ballast and sleeper surfaces, the absolute offset does not influence the subsequent geometric analysis.

By removing only the spatial distortion while preserving the global depth level, the corrected depth map faithfully reflects the true relative geometry of the scene. This stabilized depth field provides a reliable foundation for constructing sleeper reference planes and performing ballast sufficiency classification in later stages.

\subsection{Plane Reconstruction and Ballast Sufficiency Classification}\label{subsec: criteria}

With the corrected depth map $ D_{\text{corr}}(x, y) $ for all $(x, y) \in \Omega$, we reconstruct the ideal sleeper-aligned reference plane for each ballast region and evaluate ballast sufficiency based on depth deviations relative to this plane. This section describes how the reference depth surface is derived within each rotated bounding box and how ballast condition is subsequently classified using robust geometric metrics.

\subsubsection{Reference Plane Construction}\label{AA}

For each rotated bounding box $\boldsymbol{b}_i^{\,r}$, we operate in its local rotated coordinate system $(u,v)$, where $u$ and $v$ denote the horizontal and vertical axes of the rotated rectangle, respectively. Let $\mathcal{R}_i^{\,r}$ denote the set of pixels belonging to the $i$-th rotated box expressed in this local coordinate frame.

Under sufficient ballast conditions, the ballast surface between the upper and lower sleeper boundaries is expected to vary approximately linearly along the $v$-axis. We therefore extract corrected depth profiles along the top and bottom edges of the rotated box. Let $z_{\mathrm{top}}(u)$ and $z_{\mathrm{bot}}(u)$ denote the depth profiles sampled at $v=0$ and $v=h_i^{r}$, where $h_i^{r}$ is the height of $\boldsymbol{b}_i^{\,r}$ in the rotated coordinate system.

For any pixel $(u,v)\in\mathcal{R}_i^{\,r}$, the reference depth is computed by linear interpolation between these two boundary profiles:
\begin{equation}
\Pi_i(u,v) = \left(1-\frac{v}{h_i^{r}}\right) z_{\mathrm{top}}(u) + \frac{v}{h_i^{r}}\, z_{\mathrm{bot}}(u).
\end{equation}

This construction yields a locally planar sleeper-aligned reference surface within each rotated box, adapting to variations in sleeper tilt and depth bias.

\subsubsection{Depth Residual Computation}\label{AA}

Given the rotated bounding region $\mathcal{R}_i^{\,r}$, the depth residual for each pixel $(u,v)\in\mathcal{R}_i^{\,r}$ is defined as
\begin{equation}
\Delta_i(u,v)
=
D_{\mathrm{corr}}(u,v)
-
\Pi_i(u,v),
\label{eq:depth_residual}
\end{equation}
where $D_{\mathrm{corr}}$ is the temporally stabilized, bias-corrected depth map expressed in the rotated coordinate frame, and $\Pi_i(u,v)$ is the reference plane derived above. Negative residuals indicate that the ballast lies below the expected reference surface and may signal insufficient material.

\subsubsection{Dual Ballast Sufficiency Classification}\label{AA}

After obtaining the residual depth map for each ballast region, we classify ballast sufficiency using two complementary criteria that correspond to the two dominant failure modes: wide-area ballast depression and localized sleeper–edge gaps. This dual-criteria design combines region-level depth statistics with boundary-sensitive analysis.

\textbf{Criterion-1 Global Residual Criterion:} To detect widespread depression, we compute the area proportion of pixels in $\mathcal{R}_i$ whose residual lies below the reference plane by more than a threshold $T_z$:
\begin{equation}
\rho_i
=
\frac{
\big\lvert
\left\{
(x,y) \in \mathcal{R}_i :
\Delta_i(x,y) < -T_z
\right\}
\big\rvert
}{
\lvert \mathcal{R}_i \rvert
},
\label{eq:global_residual}
\end{equation}
where $\lvert \cdot \rvert$ denotes the number of pixels in each set, equivalently representing the relative area. Larger values of $\rho_i$ indicate that a significant portion of the ballast region is depressed.

\textbf{Criterion-2 Edge Gap Criterion:} Localized ballast loss often appears near the sleeper–ballast interfaces. For a rotated bounding box $\boldsymbol{b}_i^{\,r}$, we define a local coordinate system $(u,v)$ aligned with the box, where $u$ denotes the horizontal axis of the rotated box and $v$ its vertical axis. Let $h_i^{r}$ denote the box height along the $v$-axis. The edge-band thickness is set adaptively as
\begin{equation}
h_{\text{edge}}^i = \kappa \, h_i^{r}, \qquad \kappa \in (0,0.5).
\end{equation}

Using this coordinate system, the top and bottom edge-band regions are defined as
\begin{equation}
\mathcal{E}_i^{\text{top}}
=
\left\{
(u,v)\in \mathcal{R}_i^{\,r} :
0 \le v < h_{\text{edge}}^i
\right\},
\end{equation}
\begin{equation}
\mathcal{E}_i^{\text{bot}}
=
\left\{
(u,v)\in \mathcal{R}_i^{\,r} :
h_i^{r}-h_{\text{edge}}^i < v \le h_i^{r}
\right\}.
\end{equation}

Let $\{u_j\}_{j=1}^{W_i}$ be the ordered set of distinct $u$-coordinates (columns). For the $j$-th column, the corresponding edge-band area is
\begin{equation}
\mathcal{E}_{i,j}
=
\left\{
(u,v) \in \mathcal{E}_i^{\text{top}} \cup \mathcal{E}_i^{\text{bot}} :
u = u_j
\right\}.
\end{equation}

The area proportion of depressed points within this column-wise edge band is defined as
\begin{equation}
\gamma_{i,j}
=
\frac{
\big\lvert
\left\{
(u,v)\in \mathcal{E}_{i,j} :
\Delta_i(u,v) < -T_z
\right\}
\big\rvert
}{
\lvert \mathcal{E}_{i,j} \rvert
},
\label{eq:edge_gap_criterion}
\end{equation}
where pixel cardinality is used as a discrete proxy for area.

Region $i$ is classified as insufficient if either criterion is activated:
\begin{equation}
s_i =
\begin{cases}
\text{insufficient}, & \text{if } \rho_i > \eta_1 \;\; \text{or} \;\;
\gamma_{i,j} > \eta_2, \\
\text{sufficient},   & \text{otherwise}.
\end{cases}
\end{equation}
Here, $\eta_1$ controls the tolerance for wide-area depression, whereas $\eta_2$ governs the sensitivity to localized edge gaps.

\section{Experiments and Results}
\label{sec:experiments}
To evaluate the proposed method, we tested it on real railroad data collected using the Intel RealSense system. Since our approach incorporates both corrected depth  (CD) and YOLO–SAM2-based rotated bounding boxes (YOLO-SAM2-RBB), we compare it with two baselines: a YOLO-only model and a YOLO–SAM2 model using axis-aligned bounding boxes (YOLO-SAM2-AABB). 
  
\subsection{Dataset Collection and Preparation}
The dataset used in this study was collected using an Intel RealSense D435 camera positioned above the railroad track, providing an orthogonal top-down view of the ballast and rail structure. The D435 recorded synchronized RGB and depth frames as the camera moved along the track. From this dataset, 1,405 images were used for training the YOLO model responsible for detecting the ballast area. An additional 418 images were reserved for evaluating the algorithm. For the training data set, the set was split into 40\% insufficient and 60\% sufficient. For the testing data set, the split was 55\% sufficient, and 45\% insufficient.
Ground-truth labels (“sufficient” or “insufficient”) were created from RGB frames extracted from the original RealSense recordings and annotated in Roboflow.


\subsection{Evaluation Metrics}
%

The Performance comparisons are evaluated using precision ($P$), recall ($R$), and F1-score ($F1$), with F1-score as the primary metric due to class imbalance. For safety-critical inspection, false positives (predicting sufficient when the region is insufficient) are the most dangerous; therefore, a higher recall for insufficient ballast is emphasized \cite{b6}.

\subsection{YOLO-Only Baseline Results}

Initial experiments evaluated the performance of the YOLOv8 model operating solely on RGB imagery without depth information. Although a high precision is obtained ($P = 0.9896$), a relatively low F1-score ($F1 = 0.6620$) is obtained due to the poor recall ($R = 0.4974$), as shown in Table I. The YOLO-Only approach tends to overpredict the “sufficient” class, leading to a high false-positive rate. YOLO-only rarely predicts ‘insufficient’, so its predictions are almost always correct when it does, but it misses many insufficient cases.

%

\subsection{Effectiveness of Depth Correction}
\begin{figure}
    \centering
    \includegraphics[width=\linewidth]{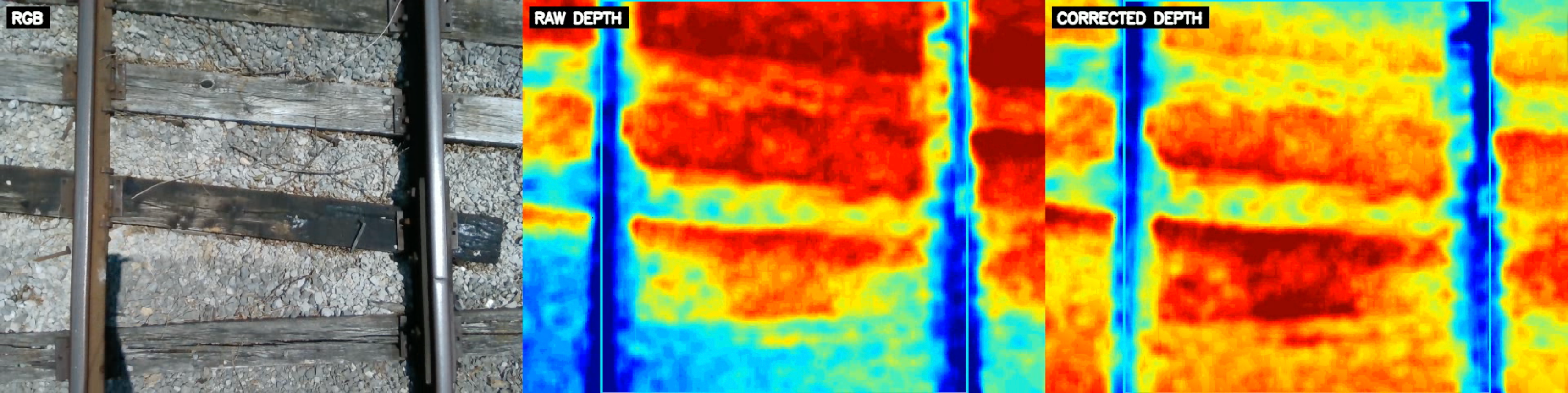}
    \caption{Comparison of RGB image (left), raw RealSense depth map (middle), and depth map after tilt and bias correction (right).}
    \label{fig:correct_depth}
\end{figure}
To address the limitations of RGB-only classification, the depth-enhanced algorithm was evaluated on the same test dataset. Fig. \ref{fig:correct_depth} illustrates the RGB frame alongside the raw and corrected depth maps used in this analysis. The corrected depth map exhibits significantly reduced spatial distortion, clearer sleeper boundaries, and a more consistent depth gradient across ballast regions, enabling more reliable detection of insufficient ballast. 


%

\subsection{Effectiveness of Rotated Bounding Boxes}
An example of a comparison between axis-aligned bounding boxes (AABB) and corresponding rotated bounding boxes (RBB) is shown in Fig. 4, which validates the effectiveness of the rotated bounding boxes method described in Section \ref{subsec: RBB}.

\begin{figure}[h!]
	\centering
	\includegraphics[width=1\linewidth]{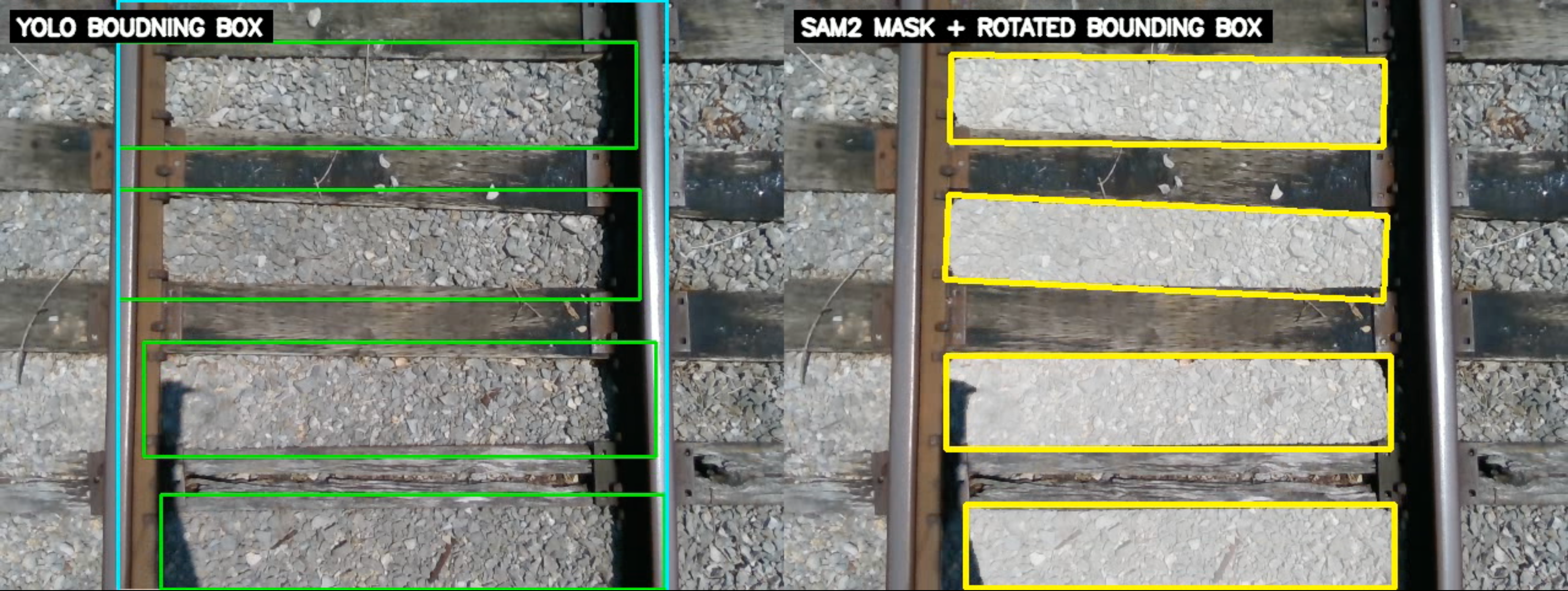}
	\caption{An example of axis align bounding boxes (left) and rotated bounding boxes (right).}
	\label{fig:RotatedBox}
\end{figure}

\subsection{Numerical Comparisons of Variant Algorithms}
To ensure consistent evaluation across all variants, we use a unified set of implementation parameters. For YOLO inference, the confidence threshold is $T_c = 0.3$ and the Intersection-over-Union (IoU) threshold used in Non-Maximum Suppression (NMS) was $0.35$. Sleeper depth samples are filtered using a MAD outlier threshold of $\tau = 3.5$. The polynomial bias surface is estimated with 160 RANSAC iterations and an inlier residual threshold of $T_{\mathrm{res}} = 0.01$ m, followed by temporal smoothing using an exponential moving average with $\lambda = 0.2$. For geometric classification, depth depressions are detected using $T_z = 0.03$ m. The global residual criterion is triggered when more than $\eta_1 = 0.4$ of the region falls below this threshold. The edge-gap criterion uses an edge-band thickness of $\kappa = 0.4$, and a column is marked insufficient when its depressed proportion exceeds $\eta_2 = 0.18$. 

To distinguish the contributions of the proposed different components, including two types of bounding boxes (presented in Section \ref{subsec: RBB}), and the different classification criteria (presented in Section \ref{subsec: criteria}), we test the variant algorithms with different components on the same image data set. The numerical comparisons are tabulated in Table \ref{tab:model_comparison}. In detail, ``CD-YOLO-SAM2" indicates the proposed approaches, ``AABB" and ``RBB" indicate the types of bounding boxes, and  ``C1" and ``C2"  indicate \textbf{Criterion-1} and \textbf{Criterion-2} in Section \ref{subsec: criteria}, respectively. Furthermore, to leverage the YOLO algorithm, we also add the decision from the YOLO method as the third criterion, denoted by ``CY" in Table \ref{tab:model_comparison}.  The decisions from all criteria, including C1, C2, and CY, are combined using a logical OR rule to generate the final decision on ``insufficient". 

From Table \ref{tab:model_comparison}, it can be observed that we can obtain the best performance when three decision criteria are all applied. In addition, we can obtain better precision performances when the RBB is applied. In contrast, we can obtain better recall performances when the AABB is applied. In sum, our proposed approaches can obtain the good precision (close to YOLO-Only), and best recall and F1-score performances. 

%

\begin{table}[htbp]
\caption{Performance Comparison of Detection Methods}
\label{tab:model_comparison}
\centering
\renewcommand{\arraystretch}{1.15}
\begin{tabular}{lccc}
\hline
\textbf{Method} & \textbf{Precision} & \textbf{Recall} & \textbf{F1} \\
\hline
YOLO Only & \textbf{0.9896} & 0.4947 & 0.6620 \\
\hline
CD-YOLO-SAM2-AABB-C1-C2 & 0.6791 & 0.7644 & 0.7192 \\
CD-YOLO-SAM2-RBB-C1-C2  & 0.7377 & 0.7068 & 0.7219 \\
CD-YOLO-SAM2-AABB-C1-CY & 0.7450 & 0.7801 & 0.7621 \\
CD-YOLO-SAM2-RBB-C1-CY  & 0.8000 & 0.7330 & 0.7650 \\
CD-YOLO-SAM2-AABB-C1-C2-CY & 0.8191 & \textbf{0.8063} & \textbf{0.8127} \\
CD-YOLO-SAM2-RBB-C1-C2-CY  & \textbf{0.8623} & 0.7539 & 0.8045 \\
\hline
\end{tabular}
\end{table}

\section{Conclusion}
This paper presented a depth-corrected YOLO–SAM2 framework with rotated bounding boxes for reliable detection of ballast insufficiency in real railroad environments. While YOLO provides strong object-level cues, our results show that RGB-only detection tends to over-predict the sufficient class, leading to poor recall for insufficient ballast—a critical limitation for safety-focused inspection. By integrating sleeper-aligned depth correction and SAM2-guided rotated bounding boxes, the proposed method significantly improves geometric alignment and reduces spatial distortion in the depth map.

Experiments on field-collected track data demonstrate that the whole system achieves the highest recall and F1-score among all tested configurations, correctly identifying insufficient ballast even in challenging conditions where RGB cues are weak. The combined effect of depth correction and rotated bounding boxes yields more stable depth profiles between sleepers, enabling more accurate geometric classification and reducing missed detections in safety-critical regions.

Future work will extend this framework to broader track conditions, incorporate multi-camera fusion for handling extreme curvature, and explore temporal consistency models for continuous, train-mounted inspection. The proposed approach provides a practical and reliable step toward automated, vision-based railway maintenance systems.

\bibliographystyle{IEEEtran}
\bibliography{references}

\end{document}